\documentclass[letterpaper]{article} % DO NOT CHANGE THIS
\usepackage[]{aaai2026}  % DO NOT CHANGE THIS
\usepackage{times}  % DO NOT CHANGE THIS
\usepackage{helvet}  % DO NOT CHANGE THIS
\usepackage{courier}  % DO NOT CHANGE THIS
\usepackage[hyphens]{url}  % DO NOT CHANGE THIS
\usepackage{graphicx} % DO NOT CHANGE THIS
\usepackage{natbib}  % DO NOT CHANGE THIS AND DO NOT ADD ANY OPTIONS TO IT
\usepackage{caption} % DO NOT CHANGE THIS AND DO NOT ADD ANY OPTIONS TO IT
\usepackage{algorithm}
\usepackage{algorithmic}

\usepackage{microtype}
\usepackage{graphicx}
\usepackage{subfigure}
\usepackage{booktabs}
\usepackage{amsmath}
\usepackage{amssymb}
\usepackage{mathtools}
\usepackage{amsthm}
\usepackage{multirow}
\usepackage{multicol}
\usepackage{longtable}
\usepackage[table]{xcolor}
\usepackage{hhline}
\usepackage[multiple]{footmisc}
\usepackage{numprint}
\npdecimalsign{\ensuremath{.}}
\newcommand{\np}[1]{\numprint{#1}}
\DeclareMathOperator*{\AGG}{AGG}
\usepackage{newfloat}
\usepackage{listings}
\DeclareCaptionStyle{ruled}{labelfont=normalfont,labelsep=colon,strut=off} % DO NOT CHANGE THIS
\floatstyle{ruled}
\newfloat{listing}{tb}{lst}{}
\floatname{listing}{Listing}
\nocopyright

\title{GravityGraphSAGE: Link Prediction in Directed Attributed Graphs}

\author{
Riccardo Porcedda\textsuperscript{\rm 1,\rm 2},
Francesca Chiaromonte\textsuperscript{\rm 1,\rm 3},
Fabrizio Lillo\textsuperscript{\rm 4},
Andrea Vandin\textsuperscript{\rm 1,\rm 5}
}
\affiliations{
\textsuperscript{\rm 1}Department of Excellence L'EMbeDS, Sant'Anna School of Advanced Studies, Pisa, Italy\\
\textsuperscript{\rm 2}Department of Computer Science, University of Pisa, Italy\\
\textsuperscript{\rm 3}Department of Statistics and Huck Institutes of the Life Sciences, The Pennsylvania State University, USA\\
\textsuperscript{\rm 4}Class of Science, Scuola Normale Superiore, Pisa, Italy\\
\textsuperscript{\rm 5}DTU Technical University of Denmark, Lyngby, Denmark
}

\begin{document}

\maketitle

\begin{abstract}
Link prediction (inferring missing or future connections between nodes in a graph) is a fundamental problem in network science with widespread applications in, e.g., biological systems, recommender systems, finance and cybersecurity. The ability to accurately predict links has significant real-world applications, such as detecting fraudulent financial transactions or identifying drug-target interactions in biomedicine. Despite a rich literature, link prediction is still challenging, especially for graphs enriched with information on edges (direction) and nodes (attributes). In fact, research on link prediction, especially the one based on Graph Deep Learning (GDL), has mostly focused on undirected graphs, without fully leveraging node attributes. Here, we fill this gap by proposing GravityGraphSAGE (GG-SAGE), a modified version of GraphSAGE, a GDL model for node embeddings, composed of a \emph{gravity-inspired decoder}. This implementation is the first example in the literature of a GraphSAGE backbone adopted for directed link prediction. Using the benchmark datasets Cora, Citeseer, PubMed and 16 real-world graphs from the online Netzschleuder repository, we show that our proposed model outperforms state-of-the-art GDL link prediction techniques. Using further experimental evidence, we relate the quality of the output of our model with various characteristics of the graph, suggesting that our framework scales well when applied to data of increasing complexity.
\end{abstract}

\begin{links}
\link{Code}{https://doi.org/10.5281/zenodo.16722908}
\end{links}

\section{Introduction}
\label{introduction}

Link prediction in complex networks is the task of inferring potential or missing links (edges) between nodes based on intrinsic features of a network \cite{Liben-Nowell}. This problem has drawn extensive attention across fields, driven by its
%applicability 
ubiquitous applications in social networks and biological systems~\cite{almansoori2012link}, recommender systems~\cite{10.1145/1065385.1065415}, finance~\cite{10.1093/jrsssa/qnad088} and cybersecurity~\cite{9832800}.

Traditional approaches 
%to 
for link prediction often rely on node similarity indices \cite{L__2011}, such as common neighbors, Adamic-Adar, and preferential attachment, which use local or global topological features to estimate the likelihood of links between node pairs. These similarity-based methods, while intuitive and computationally feasible, are typically limited to undirected graphs and struggle with more complex directed, attributed graphs.
%These approaches were then outperformed when 
Moreover, they were outperformed by the introduction of Deep Learning methods 
%have been extended to 
for graph data, 
%leading to 
which led to Graph Deep Learning \cite{9039675}. 
%This was obtained first by using
The first proposals, where graph measures were used as input 
%data 
for basic neural networks \cite{zhang2016deep}, 
%and later by developing 
were soon followed by the development of Graph Neural Networks (GNNs), 
%that could learn directly from the network as input data (
which directly take as input networks represented 
%with 
through adjacency or laplacian matrices~\cite{DBLP:journals/corr/DefferrardBV16}.
The general framework of GNNs for link prediction is the following~\cite{NEURIPS2018_53f0d7c5}: first, train the model to learn a low-dimensional vector representation for the nodes, i.e.~an
%the 
\emph{embedding} (this 
%could be done both in an unsupervised fashion or a supervised one
can be done either in an unsupervised or in a supervised fashion);
%, 
then, predict links by using a proximity measure 
%(like inner product of two vectors) 
(e.g., the inner product of two vectors) in the embedding space.
However, and notably, most 
%of these 
GNN advances to date
%concerned only 
concern undirected networks without attributes on nodes. 

\paragraph{Contribution.}
In this work, we 
%solve this problem 
fill this gap by proposing a GNN approach for link prediction in directed, attributed graphs.
Our proposal consists of combining a variant of GraphSAGE equipped with an ELU activation function on the final embedding layer, with a \emph{gravity-inspired} decoder that was first proposed in \cite{salha2019}, but never applied to this architecture.
The former component 
%helps for handling 
allows us to handle node attributes in large-scale networks, while the latter accounts for edge directionality.
Using real-world graphs from 
%an 
the Netzschleuder online repository, we 
%are able to 
show that 
%this framework 
our proposal outperforms state-of-the-art link prediction techniques.

\paragraph{Synopsis.}
Section~\ref{sec: preliminaries} introduces some taxonomy in order to better define node embeddings (what they are, how they are generated, which are the state-of-the-art models) and the challenges posed by directed graphs and link prediction in GDL.
Section~\ref{sec: gravityGraphSAGE} provides details of our 
%approach
proposal, the GravityGraphSAGE model, outlining how it differs from prior work by combining the inductive power of GraphSAGE with the gravity-inspired decoder.
Section \ref{sec: experiments} describes the experimental settings and compares our model with state-of-the-art GDL methods 
%on 
on directed and attributed 
%datasets 
graphs selected from the Netzschleuder repository (\url{https://networks.skewed.de/}).
Finally, Section~\ref{sec: conclusions} 
%concludes the paper 
provides final remarks and outlines 
%diections 
directions for future research.

%\section{Node Embeddings} 
%\section{\coma{Related works}} 
\section{Preliminaries and related works}
\label{sec: preliminaries}

We now provide 
%the necessary 
relevant preliminary material and highlight related works on which our method is based.

\subsection{Directed, unweighted, and attributed graphs}
\begin{figure}[h]
    \centering
    \includegraphics[width=0.5\linewidth]{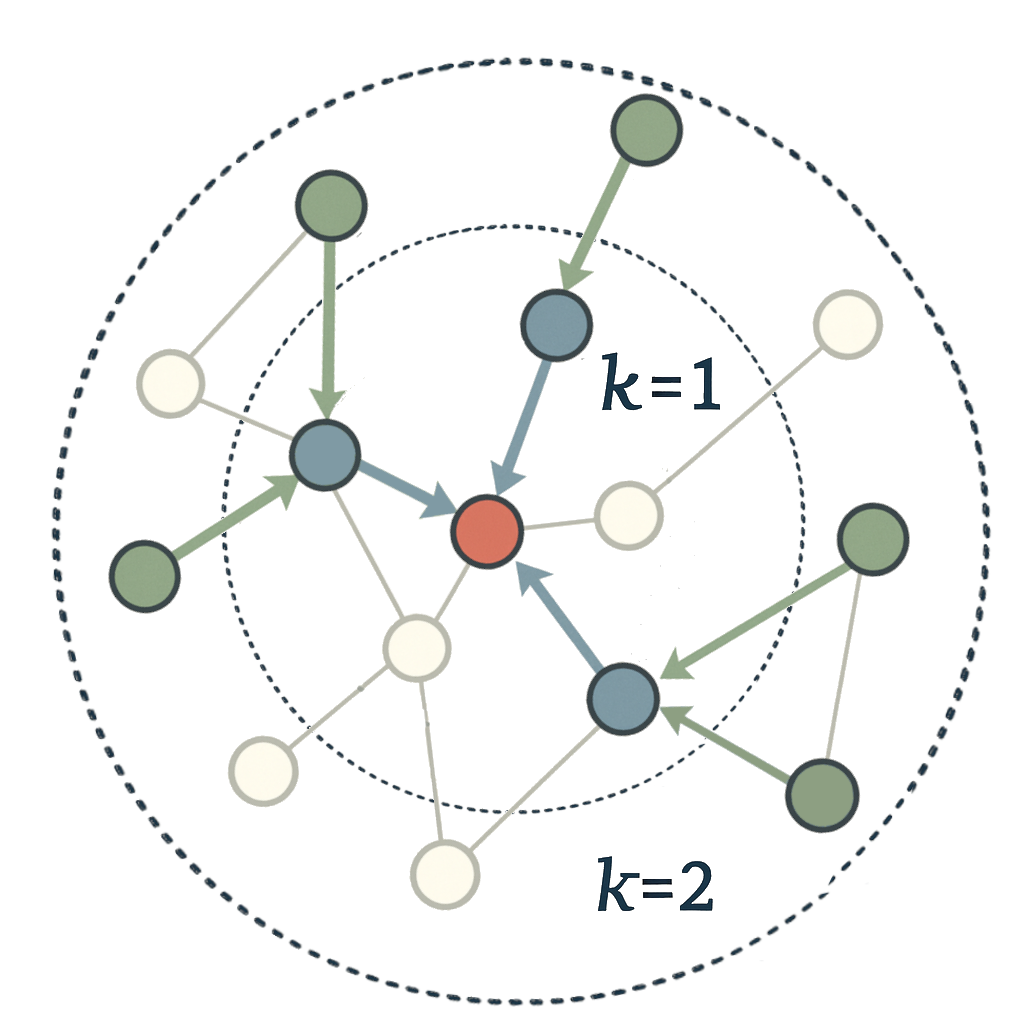}
    \caption{Example of K-hop %neighbourhood
    neighborhood}
    \label{fig:k-hop}
\end{figure}

Let $G(\mathcal{V}, \mathcal{E}, F)$ be a directed, unweighted and attributed graph, where $\mathcal{V}$ represents the set of $n$ nodes, $\mathcal{E} \subseteq \mathcal{V} \times \mathcal{V}$ is the set of $m$ edges and $F \in \mathbb{R}^{n \times d}$ is the node attribute matrix with $d$ being the number of features. An edge $e_{ij} \in \mathcal{E}$ connects node $v_i$ to node $v_j$. A graph can also be compactly represented as an adjacency matrix $A$ given as $A_{ij}=e_{ij}$ for all $i,j$. 
%\coma{Less is more. Ci serve davvero $\mathcal{E}$? Non possiamo fare subito con la adjacency matrix?}
In the case of a directed and unweighted graph, this matrix is binary, $A_{ij} \in \{0,1\}$. %, and not necessarsymmetric, $A \neq A^T$.\coma{leva not symmetric. 1to2,2to1 lo e'. Non lo e' per forza, ma puo' esserlo}
The shortest distance from node $i$ to node $j$ is the minimum number of edges that have to be crossed to get from $i$ to  $j$. 
As shown in Figure~\ref{fig:k-hop}, the K-hop 
%neighbourhood 
neighborhood $\mathcal{N}_K(v)$ of a node $v$ is the set of nodes at distances at most $K$ from that node.

In terms of adjacency matrix, we have
%: 
$\mathcal{N}_K(v)=\{u|A^K_{uv} \neq 0\}$. If we consider the whole 
%neighbourhood
neighborhood, 
%i.e., 
without posing a bound 
%on 
$K$ on the distance,
%is considered, 
then we write $\mathcal{N}(v)$.

%\subsection{Overview of node embedding techniques}
\subsection{Node Embeddings}
\label{subsec: node_embeddings}
Node embeddings, i.e. low-dimensional vector representations of nodes in a graph, are a fundamental technique in graph-based machine learning and are employed in various downstream tasks, including node classification, link prediction, and clustering \cite{8294302}.
Traditional methods, such as matrix factorization \cite{Belkin-Niyogi, HOPE}, laid the groundwork for embedding techniques by decomposing adjacency or Laplacian matrices to capture latent relationships. However, these approaches faced challenges in scalability and flexibility. The advent of random walk-based methods, like DeepWalk \cite{DeepWalk} and node2vec \cite{grover2016node2vecscalablefeaturelearning}, extended neural language models to graph structures to learn node representations, effectively capturing both local and global structural information.
%
%Still, 
Notably though, it has been shown that these models, too, rely on some kind of matrix factorization \cite{NIPS2014_feab05aa, Network_Embedding_as_Matrix_Factorization}.

\begin{figure}[b]
    \centering
    \includegraphics[width=0.5\linewidth]{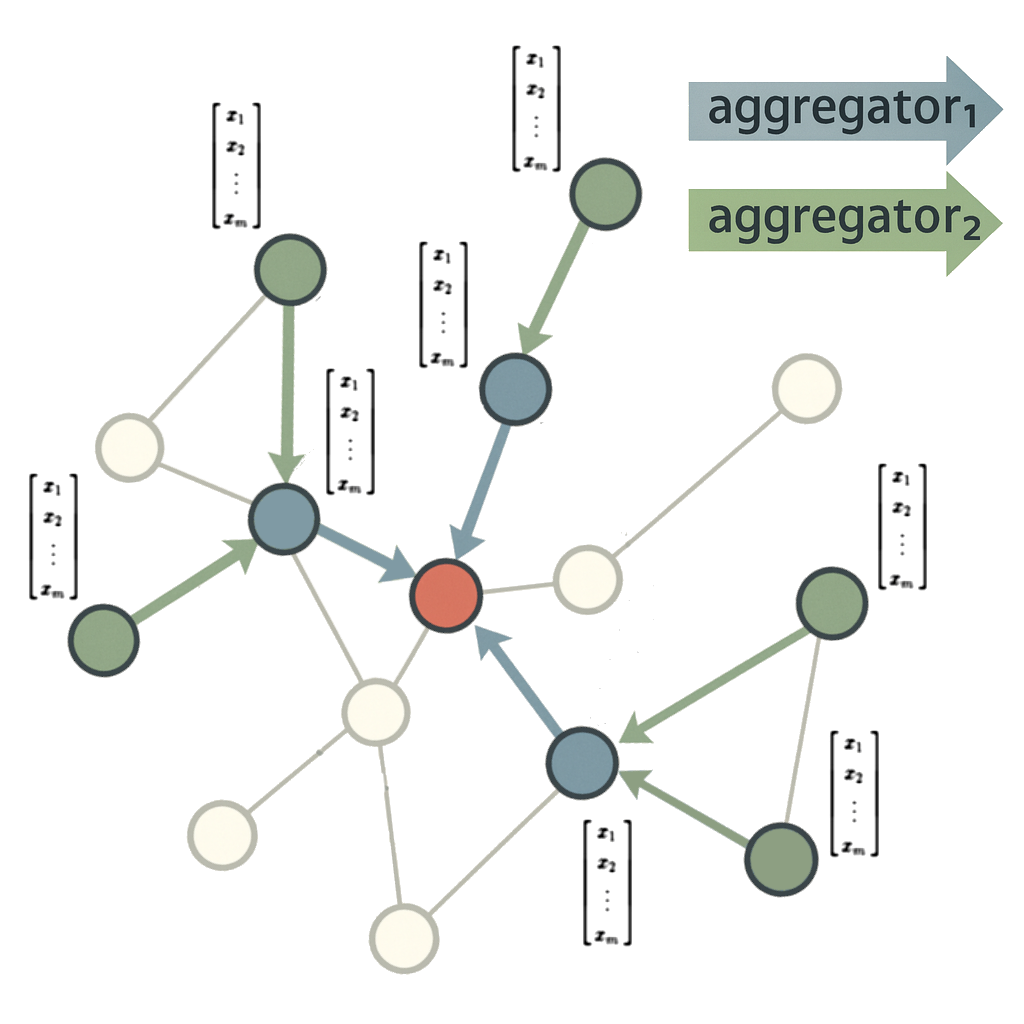}
    \caption{Example
    %Illustrative example 
    of message-passing}
    \label{fig:message-passing}
\end{figure}

Later, more complex GNNs have been developed to leverage both node attributes and topological information.
The 
%common 
main idea 
%of 
shared by these models is,
%that, 
for each node, to aggregate the features of its 
%neighbours 
neighbors 
%are aggregated, and then combined to its own features
and combine them to the features of the node itself. This process is referred to as \emph{message-passing} (Figure \ref{fig:message-passing}).

%Different 
Message-passing 
%forms exist. 
can be implemented in different ways. The most popular 
%examples are 
utilize architectures such as Graph Convolutional Networks (GCN) \cite{kipf2017semisupervised}, Graph Attention Networks (GAT) \cite{veličković2018graph} and GraphSAGE \cite{hamilton2018inductiverepresentationlearninglarge}.
The key features of these 
%three 
architectures are summarized in Table~\ref{tab:gnn_comparison}
%, 
and further discussed below. 
%in what follows.

\begin{table}[t]
    \centering
    \begin{tabular}{l p{5.4cm}} % Adjust column width as needed
        \toprule
        \textbf{Model} & \textbf{Description} \\
        \midrule
        \textbf{GCN} & \textbf{Key Idea:} Aggregates and averages neighbor features via message passing \newline 
                     \textbf{Strengths:} Efficient due to sparse matrix multiplication \newline
                     \textbf{Weaknesses:} Assigns equal importance to all neighbors; requires full adjacency matrix, limiting scalability \\
        \midrule
        \textbf{GAT} & \textbf{Key Idea:} Uses \textit{self-attention} to assign adaptive weights to neighboring nodes \newline 
                     \textbf{Strengths:} Can learn specific importance of neighbors\newline
                     \textbf{Weaknesses:} Computationally expensive due to pairwise attention computations; requires full adjacency matrix \\
        \midrule
        \textbf{GraphSAGE} & \textbf{Key Idea:} Uses \textit{sampling} to aggregate neighborhood information instead of full adjacency matrix \newline 
                            \textbf{Strengths:} Can generalize to unseen nodes; it does not require full graph storage; better handles oversmoothing \newline
                            \textbf{Weaknesses:} Sampling introduces stochasticity in training; ignores \textit{distant neighbors} beyond sampled depth \\
        \bottomrule
    \end{tabular}
    \caption{Comparison of main GNN architectures}
    \label{tab:gnn_comparison}
\end{table}

From now on, let 
%be:
\begin{itemize}
    \item $ h_v^{(l)} \in \mathbb{R}^{d_l}$ be the representation of node $ v $ at layer $ l $;
    \item $ W^{(l)} \in \mathbb{R}^{d_{l-1}\times d_l}$ 
    %is 
    be the learnable weight matrix; and
    \item $ f^{(k)} $ 
    %the 
    be an activation function (e.g., ReLU).
\end{itemize}

At layer $l=0$ (the input), we have $h_v^{(0)}=F_v$, i.e. the original attributes of node $v$.
%'s attributes.
%So, 
Thus, defining $H^{(l)}= [h_1^{(l)} \, h_2^{(l)} \, \ldots \, h_n^{(l)}] \in \mathbb{R}^{n\times d_l}$ for the entire graph, 
%it is clear that 
we have $H^{(0)}=F$.
If no node attribute matrix is given, then we set $F=I$.
%is set.

\paragraph{Graph Convolutional Networks (GCN).}
GCNs extend the concept of convolution from grid-like structures (e.g., images) to graphs.
The message-passing form here is
%:
\begin{equation*}
    h_v^{(l)} = f^{(l)} \left( W^{(l)} \cdot \frac{h_v^{(l-1)} + \sum\limits_{u \in \mathcal{N}(v)} h_u^{(l-1)}}{1+|\mathcal{N}(v)|} \right)
\end{equation*}
which can 
%also 
be rewritten 
%in a more compact way:
more compactly as
\begin{equation*}
    H^{(l)} = f^{(l)}(D^{-1} A H^{(l)} W^{(l)})
\end{equation*}
%with 
where $D \in \mathbb{R}^{n\times n}$ is the diagonal node degree matrix.

The message-passing in GCNs is basically an average of the selected node representation $h_v^{(l-1)}$ and the aggregated (summed) representations of its neighborhood.
This means that all neighbors are given
%have 
equal importance, which may not always be 
%valid
appropriate.
Furthermore, while the convolution operation in matrix form can leverage sparse matrix multiplication for scalability, it is important to note that GCNs require the entire adjacency matrix; this poses a challenge when
%. The implication being that
scaling to large graphs.
%poses a challenge.
%\begin{itemize}
    %\item 
%\paragraph{Pros.}
%GCNs are following efficient\coma{in what? can be trained efficiently? can be used efficiently?}, leveraging \textit{sparse matrix multiplications} for scalability.
    %\item 
%They use \textit{message passing} to propagate information across neighborhoods. 
    %\item 
%They perform well on \textit{homophilic graphs}, where connected nodes tend to have similar attributes.
%\end{itemize}

%\paragraph{Cons.}\coma{trasforma in test}
%\begin{itemize}
%    \item Assumes \textit{equal importance} for all neighbors, which may not always be valid.
%    \item Struggles with \textit{heterophilic graphs}, where nodes with different labels should be connected.
%    \item Requires the entire graph during training, making it \textit{less scalable} for large-scale graphs.
%\end{itemize}

\paragraph{Graph Attention Networks (GAT).}

GATs introduce \textit{self-attention mechanisms} \cite{bahdanau2014neural, vaswani2023attentionneed} to dynamically weigh neighbor contributions. The update rule here is
%:
%
\begin{equation*}
%\begin{split}
   \!\! h_v^{(l)} %&
   \!=\! f^{(l)}\!\!
    %&
    \left(
        \!\!W^{(l)} \!\! \left[
            \sum\limits_{u \in \mathcal{N}(v)} \!\!\!\!\!\alpha_{vu}^{(l-1)} h_{u}^{(l-1)}
            + \alpha_{vv}^{(l-1)} h_{v}^{(l-1)}
        \right]\!
    \right)\!\!\!\!
%\end{split}
\end{equation*}

where $ \alpha_{vu} $ is the \textit{attention coefficient},
%determining 
which represents the importance of neighbor $u$ to node $v$.

The 
%evident 
obvious advantage of this form of message-passing is that, unlike GCNs, it does not assign the same relevance to the whole
%neighbourhood, implying 
neighborhood -- thus allowing better aggregation and feature learning.
%This also has the 
The inevitable 
%drawback is
drawbacks are a larger
%of adding more 
number of parameters and
%increasing the computational demand 
an increased computational burden due to the pairwise attention 
%coefficient 
coefficients $ \alpha_{vu} $.
%\textbf{Pros:}
%\begin{itemize}
%    \item Assigns \textit{adaptive weights} to neighbors, improving expressive power.
%    \item Can handle \textit{heterophilic graphs} better than GCNs.
%    \item Multiple attention heads allow for \textit{richer feature learning}.
%\end{itemize}

%\textbf{Cons:}
%\begin{itemize}
%    \item Computationally expensive due to \textit{pairwise attention computation}.
%    \item Requires \textit{more parameters}, leading to potential overfitting.
%    \item Scaling to large graphs is challenging without efficient approximations.
%\end{itemize}

\paragraph{GraphSAGE (Sample and Aggregate).}

GCNs and GATs require access to the full graph (the entire adjacency matrix) during training.
%, and, 
Hence, if nodes are added to the graph,
%later, 
a new training is required;
%: this is what 
this is called \emph{transductive learning}.
In contrast, GraphSAGE is designed for \emph{inductive learning}, meaning that it can leverage node attribute information to efficiently generate representations on previously unseen data.
%It 
GraphSAGE's uses a neighborhood sampling strategy through the update rule
%:
%
\begin{equation} \label{eq:graphsage}
    h_v^{(l)} = f^{(l)} \left(
        W^{(l)}  \left[
            \AGG\limits_{u \in \mathcal{N}_K(v)} \left(\left\{h_{u}^{(l-1)}\right\}\right), h_{v}^{(l-1)}
        \right]
    \right)
\end{equation}
where the aggregation function can be
%mean
averaging, LSTM-based, or pooling. 
%The expression for mean aggregation is
In symbols, aggregation through averaging has the form
%:
\begin{equation}\label{eq:mean_agg}
     \AGG\limits_{u \in \mathcal{N}_K(v)} \left(\left\{h_{u}^{(l-1)}\right\}\right) = \frac{h_v^{(l-1)} + \sum\limits_{u \in \mathcal{N}_K(v)} h_u^{(l-1)}}{1 + |\mathcal{N}_K(v)|} \ .
\end{equation}
%
%It can be noticed 
Note that this aggregation function leads to a message-passing rule 
%which is 
similar to that of GCNs.
%'. But GraphSAGE, 
However, before applying the transformation $W^{(k)}$, GraphSAGE concatenates the aggregated features of the neighborhood $\mathcal{N}_K(v)$ with the features of the node $v$.
%'s features: this 
This avoids \emph{oversmoothing}
%) 
of node embeddings, 
%meaning similarity of 
i.e.~excessively similar representations 
%among 
for different nodes.
%It is in fact shown 
It has been shown that GraphSAGE, within the first three layers, tackles this problem better than GCNs and GATs \cite{Chen_Lin_Li_Li_Zhou_Sun_2020}.
%Being 
Since 
%the 
aggregation here is the result of a concatenation, for the weight matrix we trivially have $W \in \mathbb{R}^{2d_{l-1}\times d_l}$.

\subsection{Link Prediction}

Once 
%the 
embeddings have been obtained -- say $h_u$ and $h_v$
%, 
for nodes $u$ and $v$, respectively -- 
%are obtained, 
the 
%usual 
typical 
%approach for 
link prediction approach 
%involves predicting the edge by using 
leverages their inner product. An edge is predicted as
%:
\begin{equation*}
    A_{uv} = \sigma (h_u \cdot h_v)
\end{equation*}
where $\sigma$ indicates the sigmoid function.
However, this 
%method has an 
approach has the inherent drawback of lacking directionality;
%: 
due to the symmetry of the inner product, the prediction is the same for
%both 
in-edges and out-edges.
%, thereby lacking directionality.

\paragraph{Directed Link Prediction.}
%To address this issue, 
One way to overcome this drawback is the Source/Target Vector Paradigm 
%has been introduced
\cite{zhou2018}, where each node 
%is represented by 
has two separate embeddings
%: one for the node 
as a source and 
%the other 
as a target. This 
%paradigm 
allows one to use
%the application of 
the inner product 
%to 
for a source/target pair of nodes, introducing asymmetry in the link prediction process:
\begin{equation*}
    A_{uv} = \sigma (h_u^{source} \cdot h_v^{target}) \ .
\end{equation*}
A similar pathway involves using complex numbers \cite{MagNet, LightDiC}, which still results in learning two vector representations for each node.
While intuitive and straightforward, this paradigm at best doubles the number of parameters to be trained. 
%this solution requires, nonetheless, twice the number of parameters to be trained. 
A more elegant solution was found by including just one additional parameter to be learned in the embedding of each node; namely, its
%learning for each node an additional parameter, the
\emph{mass}
%, included in the embedding
\cite{salha2019}.
This approach 
%is 
was inspired by Newtonian gravity, where the force felt by two 
%masses 
bodies results in different accelerations due to the difference in their masses; in symbols
%their differences in mass:
\begin{equation*} \label{eq:gravity}
%\begin{split}
    F %&
    = G \frac{m_1 m_2}{r^2}\qquad \quad
    a_{1 \rightarrow 2} %&
    = \frac{F}{m_1} = \frac{G m_2}{r^2}
%\end{split}
\end{equation*}
%\textcolor{red}{where... [EXPLAIN THE SYMBOLS?]}
where $G$ is the gravitational constant, $m_1$, $m_2$ are the masses, $r$ is the distance between those two.
%Defining 
If one defines the embedding $h_u$ 
%for any 
of node $u$ as
\begin{equation*}
    h_u=\begin{pmatrix}
        \bar{h}_u \\
        \log(G m_u)
    \end{pmatrix}
    =\begin{pmatrix}
        \bar{h}_u \\
        \tilde{m}_u
    \end{pmatrix}
\end{equation*}
%
%and considering 
where $\bar{h}_u$ can be interpreted as
%corresponds to 
%be 
the position of mass $m_u$ in the embedding space
%\textcolor{red}{[I AM NOT SURE I UNDERSTAND THIS]}
, the acceleration in Equation~\ref{eq:gravity} can be rewritten, on the log scale, as
\begin{equation*}
\begin{split}
    A_{uv} &= \log(a_{u \rightarrow v}) \\
     &= \sigma (\log(G m_v) - \log\|h_u - h_v\|^2 )\\
        &= \sigma (\tilde{m}_v - \log\|h_u - h_v\|^2 ) \ .
\end{split}
\end{equation*}
This asymmetric quantity is then used as a prompt for directed link prediction (the logarithm is used because, transforming ratios into differences, it makes computation more stable).
Other examples of this approach include the use of a gravity decoder in a hyperbolic space \cite{D-HYPR}.
%Being this log-acceleration logarithm allows to transform the division into a subtraction, making the computation more stable) asymmetric, it is used as a prompt for directed link prediction.

\section{GravityGraphSAGE}
\label{sec: gravityGraphSAGE}

\begin{figure}[t]
    \centering
    \includegraphics[width=0.8\linewidth]{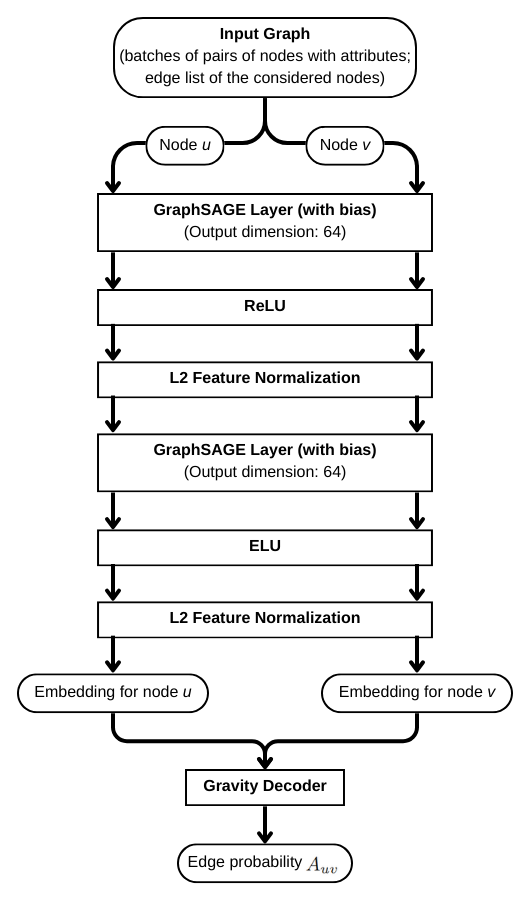}
    \caption{Schematic illustration of GG-SAGE}
    \label{fig:ggsage}
\end{figure}

%We are now in a position to present our method. 
So far we 
%have 
discussed state-of-the-art approaches for link prediction based on GNNs. In their original formulations, none of these approaches supported \emph{directed} link prediction~\cite{NEURIPS2018_53f0d7c5, 9431673, PAPADIMITRIOU20122119, DBLP:journals/corr/WangXWZ14, pan2021neural, ucar2023nessnodeembeddingsstatic}. %{\bf FAB: Perche' queste parentesi?} \textcolor{red}{[MISSING CITATIONS?]}. 
%There exist 
Hovewer, proposals have been introduced to 
%mitigate 
address this issue for two out of three of the architectures considered.
%; namely, for GCNs and GATs.
For GCNs, Salha 
%et al. 
and colleagues~\cite{salha2019} implemented a Graph AutoEncoder (GAE) and a Graph Variational AutoEncoder (GVAE). They 
%test 
tested these architectures, first integrating them with their gravity-inspired decoder (Gravity GAE/VGAE) and then following the Source/Target Vector Paradigm (Source/Target GAE/VGAE).
%
%As regards 
For GATs, there are examples in the literature 
%about the application of 
employing the Source/Target Vector Paradigm \cite{10.1093/bib/bbac151, DUPLEX}. 
%
%
%Instead, 
Notably though, no proposals exists in the literature 
%exists 
to extend GraphSAGE with directed link prediction capabilities. 

%In this paper, 
We fill this gap 
%by proposing 
introducing GravityGraphSAGE (GG-SAGE), an extension of GraphSAGE for directed link prediction. 
As 
%shown 
illustrated in Figure~\ref{fig:ggsage}, GG-SAGE uses the 
%(already discussed) 
gravity-inspired decoder discussed above. Furthermore, 
%GG-SAGE is composed of 
it comprises two message-passing layers based on
%using 
the 
%mean 
average aggregator described in Equation~\ref{eq:mean_agg}:
\begin{equation*}
    \begin{split}
        h_v^{(1)} &= ReLU \left(
        W^{(1)}  \left[
            \AGG\limits_{u \in \mathcal{N}(v)} \left(\left\{h_{u}^{(0)}\right\}\right), h_{v}^{(0)}
        \right]
        + b^{(1)}
    \right) \\
        \tilde{h}_v^{(1)} &= \frac{h_v^{(1)}}{||h_v^{(1)}||_2}
        \end{split}
\end{equation*}

\begin{equation*}
    \begin{split}
        h_v^{(2)} &= ELU \left(
        W^{(2)} \left[
            \AGG\limits_{u \in \mathcal{N}(v)} \left(\left\{\tilde{h}_{u}^{(1)}\right\}\right), \tilde{h}_{v}^{(1)}
        \right]
        + b^{(2)}
        \right)\\
        \tilde{h}_v^{(2)} &= \frac{h_v^{(2)}}{||h_v^{(2)}||_2} \ .
    \end{split}
\end{equation*}
%with the mean aggregator described in Equation \ref{eq:mean_agg}.
%
Comparing this model to the original GraphSAGE in Equation \ref{eq:graphsage}, some important choices and additions deserve a deeper discussion. 
First, 
%of all, a 
we add a bias term $b^{(l)} \in \mathbb{R}^{d_l}$. 
%has been added. Furthermore,
Second, we apply an L2 feature normalization 
%is applied 
at each layer.
%: this procedure has the effect of reducing 
This reduces oversmoothing and establishes a direct connection between node similarity and the Euclidean distance we adopt in the decoder~\cite{DBLP:journals/corr/abs-2103-00164}.
%A third noteworthy aspect is that 
Third, we use ELU \cite{clevert2015fast} as activation function for the final layer rather than ReLU. The ELU activation function has negative output values, a \textit{desideratum} in this setting for two reasons:
\begin{itemize}
    \item negative values push mean unit activations closer to zero, like batch normalization, but with lower computational complexity --
    %. It 
    this has been shown 
    %that this speeds 
    to speed up learning \cite{clevert2015fast};
    %.
    \item negative values for the mass parameter allow
    %for 
    the gravity-inspired decoder to account for repulsive effects between nodes.
\end{itemize}

In the next section we experimentally demonstrate that these components lead to a method for directed link prediction which outperforms state-of-the-art competitor techniques in terms of established metrics. 

\section{Experiments
%Experimental Analysis
} 
\label{sec: experiments}

In this section, we outline the experimental setup used to validate GG-SAGE, including details on the datasets, the implementation of the validated models, and their training configurations. We then analyze the results, providing insights into the relationship between
%models' performances 
model performance and graph properties.

\subsection{Experimental setting}
We run our experiments on a virtual machine with 40 Intel(R) Xeon(R) Gold 6252 CPUs (base frequency of 2.10 GHz) and 96 GB of RAM.
We compare GG-SAGE with Gravity GAE/VGAE, Source/Target GAE/VGAE \cite{salha2019}, LightDiC \cite{LightDiC} and D-HYPR \cite{D-HYPR} (other above-mentioned models were not considered for experiments because they are either outperformed by the chosen ones or because the replicability material is absent or not functioning).

All models are implemented using PyTorch Geometric\footnote{LightDiC: \url{https://github.com/xkLi-Allen/LightDiC}}\footnote{D-HYPR: \url{https://github.com/hongluzhou/dhypr}}. The original implementations of Gravity GAE/VGAE and Source/Target GAE/VGAE were based on Tensorflow, but the authors also referenced a PyTorch version\footnote{\url{https://github.com/ClaudMor/gravity_gae_torch_geometric}} which we include in our replicability material.
In order to ensure a fair 
%evaluation, all compared 
comparison, all architectures are constructed with two message-passing layers, and with common parameters. In particular, we use hidden dimension $d_l=64$ for both $l=1$ and $l=2$, 
%while 
and for the training routine
%, the choice fell on 
we employ a learning rate of $0.001$, $200$ epochs with early stopping and a batch size of $128$.

\subsection{Task definition}
In order to evaluate the models we perform a 5-fold cross validation with resampling. In each iteration, we split training, validation and test set as follows.
%: 
As training set we use incomplete versions of a graph where 15\% of the edges are randomly removed 
%while taking  into account 
accounting for directionality %(i.e., 
(if an edge between nodes $ u $ and $ v $ is reciprocal, we may remove the edge $ (u, v) $ while still retaining its reverse counterpart
%, 
$ (v, u) $
%, 
in the training graph).
Furthermore, as commonly done in the literature~\cite{NEURIPS2018_53f0d7c5, 9431673, PAPADIMITRIOU20122119, DBLP:journals/corr/WangXWZ14, pan2021neural, ucar2023nessnodeembeddingsstatic, salha2019}, we pair edges with an equal number of randomly sampled non-existent edges (i.e., pairs of nodes that were not originally connected).
This 
%procedure 
is commonly known as \emph{negative sampling}~\cite{10.1145/3394486.3403218}. 
%
%The 
Validation and test sets are built using the 
%removed 
edges removed from the training set; specifically
%, namely 
5\% are assigned to validation and 10\% to testing.
%, respectively. 
Again, as done for the training set and in line with
%as usual in 
the literature, we add negative samples to both validation and test sets.
%similarly to what done for the training set.
%
The validation set is used exclusively for early stopping to optimize model performance.

The performance of a model is evaluated on a binary classification task, where each pair of nodes has label 1 if connected by an edge, and 0 otherwise. %iswhere the goal is to distinguish the real (removed) edges from the negative samples. % (the ones added to the test and validation sets).

Following the literature~\cite{NEURIPS2018_53f0d7c5, 9431673, PAPADIMITRIOU20122119, DBLP:journals/corr/WangXWZ14, pan2021neural, ucar2023nessnodeembeddingsstatic, salha2019}, 
we use the AUC (Area Under the ROC Curve) and AP (Average Precision, the area under the precision-recall curve) 
%scores 
as evaluation metrics.
%for the evaluation.
AUC measures the probability that a randomly chosen positive example (a real edge) is ranked higher than a randomly chosen negative example (a non-existent edge).
Instead, AP focuses on performance for the positive class (edges that exist).
These are the two most employed metrics for classification tasks, since they are \emph{threshold-independent}
%. That is, 
(unlike accuracy, 
%AUC and AP 
they do not require choosing a classification threshold).
%for classification).

\subsection{Datasets}
For a first quick evaluation, we use the well-known Cora, Citeseer and PubMed datasets. Later on, to ensure a broad, domain-agnostic benchmarking and for the purpose of our ablation study in Section \ref{sec:ablation},
%it has been chosen to 
we use the Netzschleuder repository (\url{https://networks.skewed.de/}).
This public online repository contains 286 datasets, for a total of \np{163735} networks, of which \np{71426} undirected, \np{92309} directed, and \np{8490} bipartite.
Among the directed graphs, \np{89307} have attributes on nodes. 
We consider all graphs as unweighted, that is, we set the weight to $1$ for all edges. 
%We select a subset of directed graphs, with attributes on nodes. 
For computational feasibility, we restrict attention to
%select all 
(directed, attributed) graphs containing
%from 
between 150 and
%to 
\np{20000} nodes, and no more than \np{200000} edges.
These are listed in Table 1 of the Supplementary Material. 
%
%Before continuing with the discussion of the experiments, we highlight 
We note that the cora-1998 dataset available in the Netzschleuder repository is not the same as the above-mentioned Cora dataset 
%vastly 
broadly used in the literature~\cite{9431673, salha2019, ucar2023nessnodeembeddingsstatic}.

To select and download 
%these 
the datasets, we used %created 
%a Python tool named 
EasyNetz\footnote{\url{https://doi.org/10.5281/zenodo.14710814}}, a Python tool that
%. It 
allows the user to easily
%define 
specify criteria such as whether the graphs are directed, undirected, weighted, unweighted, and various properties like the number of nodes, edges, etc. 
%, and graph density. 
Once the graphs are filtered according to the criteria, 
%the selected graphs 
they can be downloaded and exported into CSV files.

%Before continuing with the discussion of the experiments, we highlight that the cora-1998 dataset available in the Netzschleuder repository is not the same as the well-known \emph{cora} dataset (2708 nodes, 5429 edges, 1433 features) vastly used in the \coma{literature~\cite{}.} 

%\section{Evaluation} 
%\label{evaluation}
\subsection{Results}

\begin{table}[ht]
\centering
\begin{tabular}{llcc}
\hline
\textbf{} & \textbf{Model} & \textbf{AUC (\%)} & \textbf{AP (\%)} \\
\hline
\multirow{7}{*}{\rotatebox{90}{\textbf{Citeseer}}} 
 & GG-SAGE & \cellcolor{green}\textbf{88.90 $\pm$ 1.56} & \cellcolor{green}\textbf{88.93 $\pm$ 1.57} \\
 \hhline{~---}
 & Gravity GAE & 72.57 $\pm$ 0.49 & 80.98 $\pm$ 0.29 \\
 & Gravity VGAE & 59.28 $\pm$ 1.54 & 66.22 $\pm$ 1.08 \\
 & S/T GAE & 76.06 $\pm$ 0.69 & 81.94 $\pm$ 0.72 \\
 & S/T VGAE & 69.84 $\pm$ 9.54 & 71.92 $\pm$ 10.48 \\
 & LightDiC & 86.10 $\pm$ 0.30 & 79.70 $\pm$ 0.20 \\
 & D-HYPR & 85.52 $\pm$ 1.13 & \textbf{86.34} $\pm$ \textbf{1.19} \\
\hline
\multirow{7}{*}{\rotatebox{90}{\textbf{Cora}}} 
 & GG-SAGE & \cellcolor{green}\textbf{93.61 $\pm$ 1.82} & \cellcolor{green}\textbf{93.69 $\pm$ 1.79} \\
 \hhline{~---}
 & Gravity GAE & 85.04 $\pm$ 0.19 & 88.57 $\pm$ 0.33 \\
 & Gravity VGAE & 56.28 $\pm$ 0.28 & 62.81 $\pm$ 0.55 \\
 & S/T GAE & 82.67 $\pm$ 0.69 & 85.53 $\pm$ 0.79 \\
 & S/T VGAE & 75.90 $\pm$ 1.14 & 77.20 $\pm$ 1.45 \\
 & LightDiC & 86.10 $\pm$ 0.30 & 80.50 $\pm$ 0.20 \\
 & D-HYPR & 88.22 $\pm$ 1.21 & 85.36 $\pm$ 1.34 \\
\hline
\multirow{7}{*}{\rotatebox{90}{\textbf{Pubmed}}} 
 & GG-SAGE & \cellcolor{green}\textbf{81.05 $\pm$ 1.48} & \cellcolor{green}\textbf{80.80 $\pm$ 1.13} \\
 \hhline{~---}
 & Gravity GAE & 76.17 $\pm$ 0.28 & 82.53 $\pm$ 0.13 \\
 & Gravity VGAE & 66.78 $\pm$ 0.36 & 70.53 $\pm$ 0.19 \\
 & S/T GAE & 69.80 $\pm$ 0.19 & 73.67 $\pm$ 0.37 \\
 & S/T VGAE & 66.20 $\pm$ 0.30 & 68.34 $\pm$ 0.50 \\
 & LightDiC & \textbf{79.50} $\pm$ \textbf{0.20} & 73.20 $\pm$ 0.20 \\
 & D-HYPR & \textbf{79.03} $\pm$ \textbf{1.21} & \textbf{79.81} $\pm$ \textbf{1.24} \\
\hline
\end{tabular}
\caption{AUC and AP results (\%) for Cora, Citeseer and Pubmed datasets. Best results are higlighted in green and in bold are indicated comparable performances.}
\label{tab:results_cora_citeseer_pubmed}
\end{table}

\begin{figure}[ht]
    \centering
    \includegraphics[width=1\linewidth]{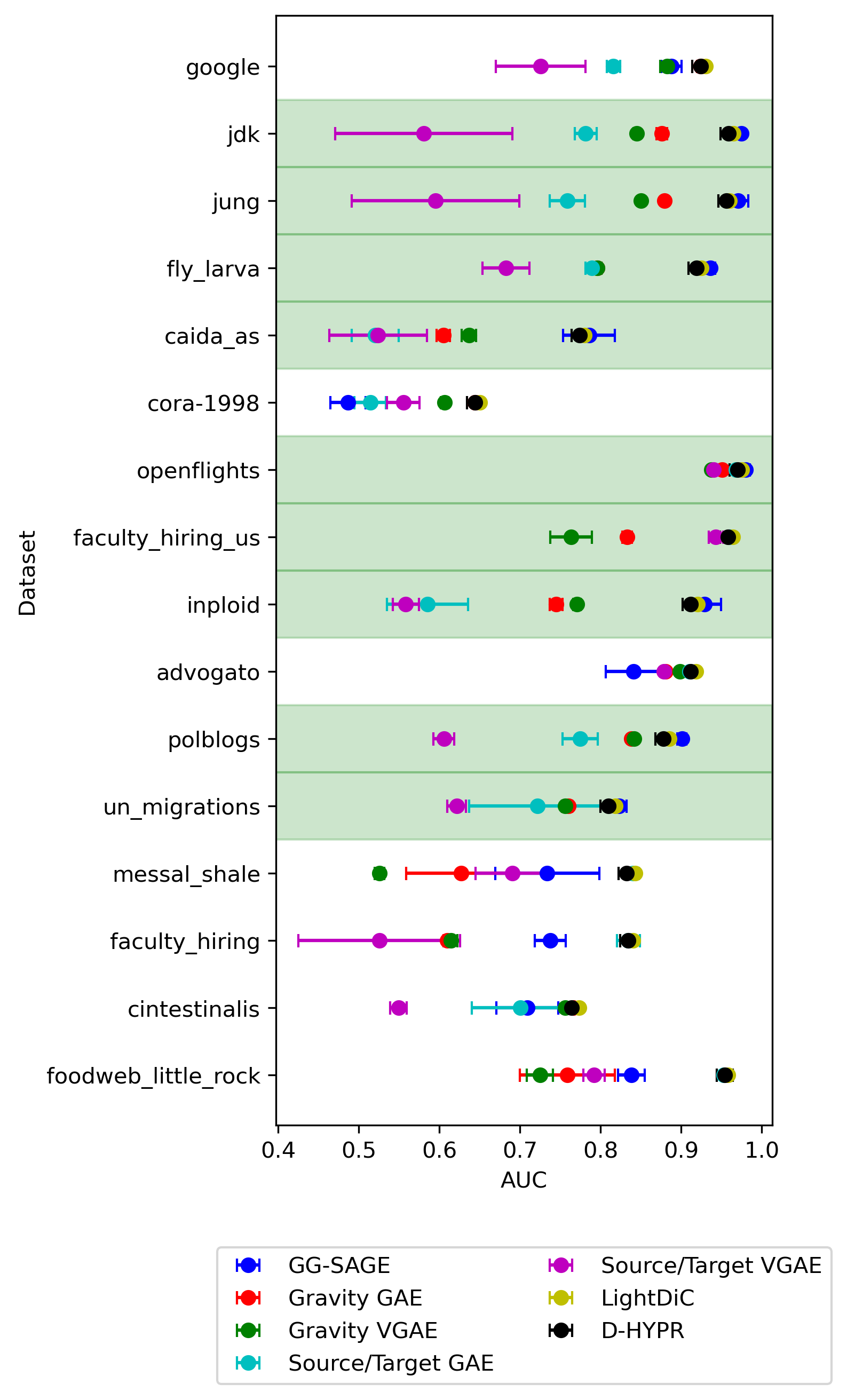}
    \caption{AUC of compared models across Netzschleuder selected datasets}
    \label{fig:auc}
\end{figure}

\begin{figure}[ht]
    \centering
    \includegraphics[width=1\linewidth]{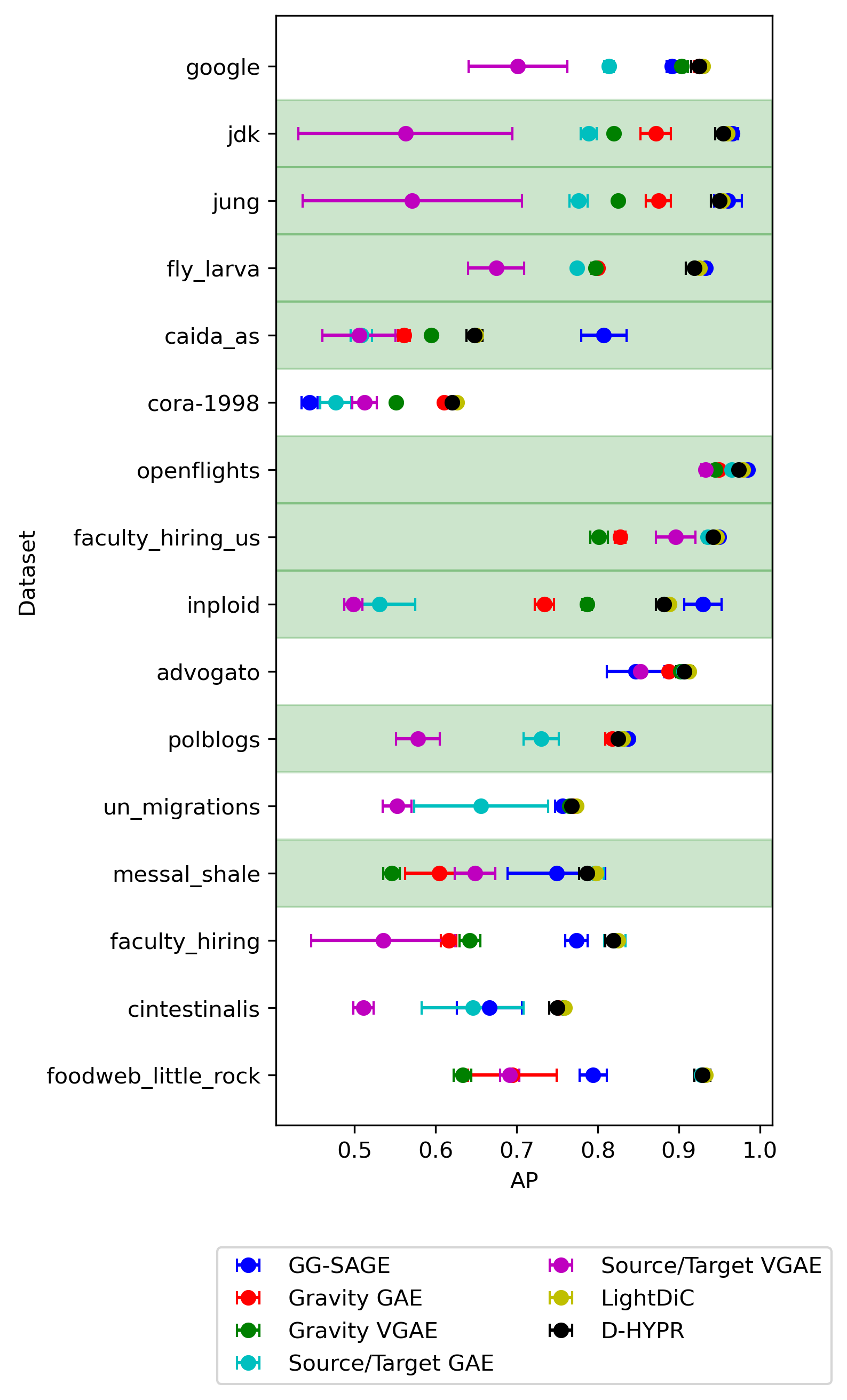}
    \caption{AP of compared models across Netzschleuder selected datasets}
    \label{fig:ap}
\end{figure}

Table \ref{tab:results_cora_citeseer_pubmed} and figures \ref{fig:auc} and \ref{fig:ap} show the results of our experiments in terms of AUC and AP, respectively. 
%We provide one line per
For each dataset (row in the figures), we report the 
%containing 
metrics as computed 
%by each considered 
for each method.
Datasets are sorted from top to bottom in order of
%according to 
decreasing size (number of edges),
%. In green, we highlight the
and datasets on which GG-SAGE
%shows 
achieves state-of-the-art
%performances 
performance are highlighted in green. In these cases (3 out of 3 for Cora, Citeseer and PubMed; 9 out of 16 Netzschleuder datasets for AUC and 10 for AP) our proposal performs best, or on par with the best performing among the considered competitors (i.e. with statistically undistinguishable metrics).
%either outperforms the competitor models or there is not statistical difference with the best performing one). This is the case in 10 out of 16 datasets.


It is evident
%immediately appears 
that GG-SAGE works best on large graphs, where it is by far the best method, with the exception of the \emph{google} and \emph{cora-1998} datasets, where the best model is LightDiC.
%Furthermore, we can notice 
There are four datasets in which LightDiC, D-HYPR and Source/Target GAE outperform by far the other models (\emph{foodweb\_little\_rock, messal\_shale, faculty\_hiring, advogato}) 
%\textcolor{red}{[WE ARE NOT ``BEST'' IN 7? OR IN 6?]}
and one in which it yields performances similar to GG-SAGE (\emph{faculty\_hiring\_us}).
Notably these datasets, which represent
%are 
food webs, trust networks, or academic flow networks,
%. These all share the fact that they are 
are all hierarchical. 
%: what these have in common, with respect to the other datasets, is that relationships in them are hierarchical.
Future work may investigate
%could try to understand better confirm and explain 
why these models better handle this type of relationships in a graph.

In Table 2 of the Supplementary Material, we report the complete numerical results for the Netzschleuder datasets, highlighting in green the best-performing model.

\subsection{Relation between
%performances 
performance and graph properties} \label{sec:ablation}
Another relevant question is how basic graph properties (number of nodes, edges and node features) influence the
%performances 
performance of the considered models.
%In order to answer 
To address this, we use a Random Forest Regression to regress 
%AUC and AP and those 
performance metrics on graph properties.
Fit results are shown in Tables \ref{tab:AUC_RF} and \ref{tab:AP_RF}.

% --------------------------  AUC  --------------------------
\begin{table}[ht]
  \centering
  \begin{tabular}{l c ccc}
    \toprule
    \multicolumn{5}{c}{\textbf{AUC – Random‑Forest Fit}}\\
    \midrule
    Model & R$^2$ & \multicolumn{3}{c}{Feature importance} \\
          &       & \#Nodes & \#Edges & \#Features \\
    \midrule
    GG-SAGE   & 0.849 & \underline{0.387} & \textbf{0.439} & 0.174 \\
    Gravity GAE       & 0.915 & 0.099 & \underline{0.238} & \textbf{0.663} \\
     Gravity VGAE      & 0.905 & 0.104 & \underline{0.244} & \textbf{0.652} \\
    S/T GAE  & 0.849 & \underline{0.386} & 0.146 & \textbf{0.468} \\
    S/T VGAE & 0.827 & 0.230 & \underline{0.275} & \textbf{0.496} \\
    LightDiC           & 0.826 & 0.270 & \underline{0.315} & \textbf{0.415} \\
    D‑HYPR             & 0.825 & 0.278 & \underline{0.319} & \textbf{0.403} \\
    \bottomrule
  \end{tabular}
  \caption{Random‑Forest $R^{2}$ and feature importances for predicting AUC. In bold, the relative importance of the most informative feature. Underlined, the second most informative.}
  \label{tab:AUC_RF}
\end{table}

% --------------------------  AP  --------------------------
\begin{table}[ht]
  \centering
  \begin{tabular}{l c ccc}
    \toprule
    \multicolumn{5}{c}{\textbf{AP – Random‑Forest Fit}}\\
    \midrule
    Model & R$^2$ & \multicolumn{3}{c}{Feature importance} \\
          &       & \#Nodes & \#Edges & \#Features \\
    \midrule
    GG-SAGE   & 0.855 & \underline{0.425} & \textbf{0.479} & 0.095 \\
    Gravity GAE       & 0.917 & 0.111 & \underline{0.269} & \textbf{0.620} \\
    Gravity VGAE      & 0.893 & 0.136 & \underline{0.312} & \textbf{0.552} \\
    S/T GAE  & 0.848 & \underline{0.333} & 0.141 & \textbf{0.526} \\
    S/T VGAE & 0.854 & 0.218 & \underline{0.248} & \textbf{0.534} \\
    LightDiC           & 0.878 & 0.197 & \underline{0.323} & \textbf{0.481} \\
    D‑HYPR             & 0.882 & 0.180 & \underline{0.318} & \textbf{0.502} \\
    \bottomrule
  \end{tabular}
  \caption{Random‑Forest $R^{2}$ and feature importances for predicting AP. Bold and underlined have the same meaning as in Table \ref{tab:AUC_RF}.}
  \label{tab:AP_RF}
\end{table}

For all models, the $R^2$ is above 85\%, hence these three variables already give a reliable, though not exhaustive, description of their behaviour.

\paragraph{One–way partial dependencies.}
In Figures 1-3 of the Supplementary Material, we visualize the marginal effect of each variable on the surrogate predictions.  
Across all methods larger edge sets correlate with higher accuracy, but the slopes differ markedly: GG-SAGE rises fastest (we notice an increment of 14\% in AUC and AP before saturating near $5 \cdot 10^{5}$ edges, whereas Gravity GAE/VGAE and Source/Target GAE/VGAE gain only $7\%$–$10\%$, while LightDiC and D-HYPR improve modestly because they already start from a high baseline.  
The dependence on the number of node features is almost flat for GG-SAGE but approximately linear for the Gravity GAE/VGAE and Source/Target GAE/VGAE, which obtain up to $+13\%$ AP when the number of features grows to $10\,000$, reaching then a plateau.  
With respect to the number of nodes, GG-SAGE, LightDiC and D-HYPR display a shallow peak around $4\,000$ nodes followed by a gradual decline, while the Gravity GAE/VGAE remain essentially level and the Source/Target variants deteriorate on large graphs.

\paragraph{Discussion.}
Taken together, the surrogate study confirms that GG-SAGE derives its predictive power primarily from topological cues—particularly edge density—while remaining largely insensitive to the variation of node attributes, but still maintaining high performances.  
All competitor models, in contrast, are feature–driven and therefore benefit strongly from high-dimensional attributes, and LightDiC as well as D-HYPR occupy a middle ground with balanced reliance on structure and attributes.  
GG-SAGE’s edge-dominated response makes it comparatively robust in scenarios where structural information is available but rich node features are sparse or noisy.

\section{Conclusions} 
\label{sec: conclusions}
In this paper, we proposed GravityGraphSAGE (GG-SAGE), a novel approach for directed link prediction in directed attributed graphs. 
%The method 
GraphSAGE is a popular graph neural network model, able to perform several tasks, which is known to 
%be able to use well
effectively use information on
%nodes 
node attributes in large graphs.
%Our proposed 
GG-SAGE 
%approach 
integrates a customized GraphSAGE backbone with a gravity-inspired decoder. To the best of our knowledge, 
%this 
ours is the first 
%application of 
GraphSAGE-based 
%approaches 
approach developed for directed link prediction.

We evaluated GG-SAGE against state-of-the-art methods for directed link prediction (LightDiC, D-HYPR, Gravity GAE/VGAE and Source/Target GAE/VGAE) on three well-known datasets (Cora, Citeseer and PubMed) and 16 real-world directed and attributed graphs from the public online repository Netzschleuder. The results demonstrate that GG-SAGE consistently outperforms, or matches, the best existing
%models 
methods, particularly excelling in large-scale networks. 

Through 
%an in-depth 
a Random Forest Regression, we showed that GG-SAGE uniquely benefits from increased graph connectivity (edge density), counteracting the performance degradation observed for all methods as the number of nodes
%grows.  
increases. This suggests that our method is particularly suited for large, well-connected graphs.
% \textcolor{red}{[DO WE NEED TO COMMENT ON THE FACT THAT GG-SAGE'S PERFORMANCE DOES NOT DEPEND ON THE NUMBER OF FEATURES? THESE ARE NODE FEATURES, CORRECT? ABOVE WE WRITE THAT GraphSAGE IS GOOD AT USING NODE ATTRIBUTES IN LARGE GRAPHS?]}

\paragraph{Future Work.}
Our study raises several interesting directions for future research. First, as discussed in Section~\ref{sec: experiments}, it is 
%interesting to confirm and understand 
of interest to investigate why the Source/Target Vector Paradigm, LightDiC and D-HYPR excel in hierarchical networks.
%; this could lead to hybrid models combining different paradigms \textcolor{red}{[I DON'T UNDERSTAND THE LAST PART]}. 
%
Another interesting aspect to
%consider is studying 
investigate is the impact of the negative sampling applied to datasets, in particular 
%in 
to the test set, commonly performed in the literature.
Lastly, it is of interest to explore 
%we will investigate 
how 
%demonstrated improved performances 
improved performance in directed link prediction, such as the one afforded by GG-SAGE, may 
%can 
benefit richer pipelines which include this step. 

%\section*{Acknowledgements}
%The work has been partially supported by 
%project SMaRT COnSTRUCT (CUP J53C24001460006), in the %context of FAIR (PE0000013, CUP B53C22003630006) under % the National Recovery and Resilience Plan (Mission 4, %Component 2, Line of Investment 1.3) funded by the %European Union - NextGenerationEU. 

\bibliography{main}


% \documentclass[letterpaper]{article} % DO NOT CHANGE THIS
% \usepackage[submission]{aaai2026}  % DO NOT CHANGE THIS
% \usepackage{times}  % DO NOT CHANGE THIS
% \usepackage{helvet}  % DO NOT CHANGE THIS
% \usepackage{courier}  % DO NOT CHANGE THIS
% \usepackage[hyphens]{url}  % DO NOT CHANGE THIS
% \usepackage{graphicx} % DO NOT CHANGE THIS
% \urlstyle{rm} % DO NOT CHANGE THIS
% \def\UrlFont{\rm}  % DO NOT CHANGE THIS
% \usepackage{natbib}  % DO NOT CHANGE THIS AND DO NOT ADD ANY OPTIONS TO IT
% \usepackage{caption} % DO NOT CHANGE THIS AND DO NOT ADD ANY OPTIONS TO IT
% \frenchspacing  % DO NOT CHANGE THIS
% \setlength{\pdfpagewidth}{8.5in} % DO NOT CHANGE THIS
% \setlength{\pdfpageheight}{11in} % DO NOT CHANGE THIS

% \usepackage{algorithm}
% \usepackage{algorithmic}

% \usepackage{microtype}
% \usepackage{graphicx}
% \usepackage{subfigure}
% \usepackage{booktabs}
% \usepackage{amsmath}
% \usepackage{amssymb}
% \usepackage{mathtools}
% \usepackage{amsthm}
% \usepackage{multirow}
% \usepackage{multicol}
% \usepackage{longtable}
% \usepackage[table]{xcolor}
% \usepackage{hhline}
% \usepackage[multiple]{footmisc}
% \usepackage{numprint}
% \npdecimalsign{\ensuremath{.}}
% \newcommand{\np}[1]{\numprint{#1}}

\title{GravityGraphSAGE: Link Prediction in Directed Attributed Graphs \\ Technical Appendix}

%\begin{document}

%\maketitle

\onecolumn

\begin{center}
    \Large \textbf{GravityGraphSAGE: Link Prediction in Directed Attributed Graphs \\ Technical Appendix}
\end{center}

\begin{figure}[h]
    \centering
    \includegraphics[width=0.9\linewidth]{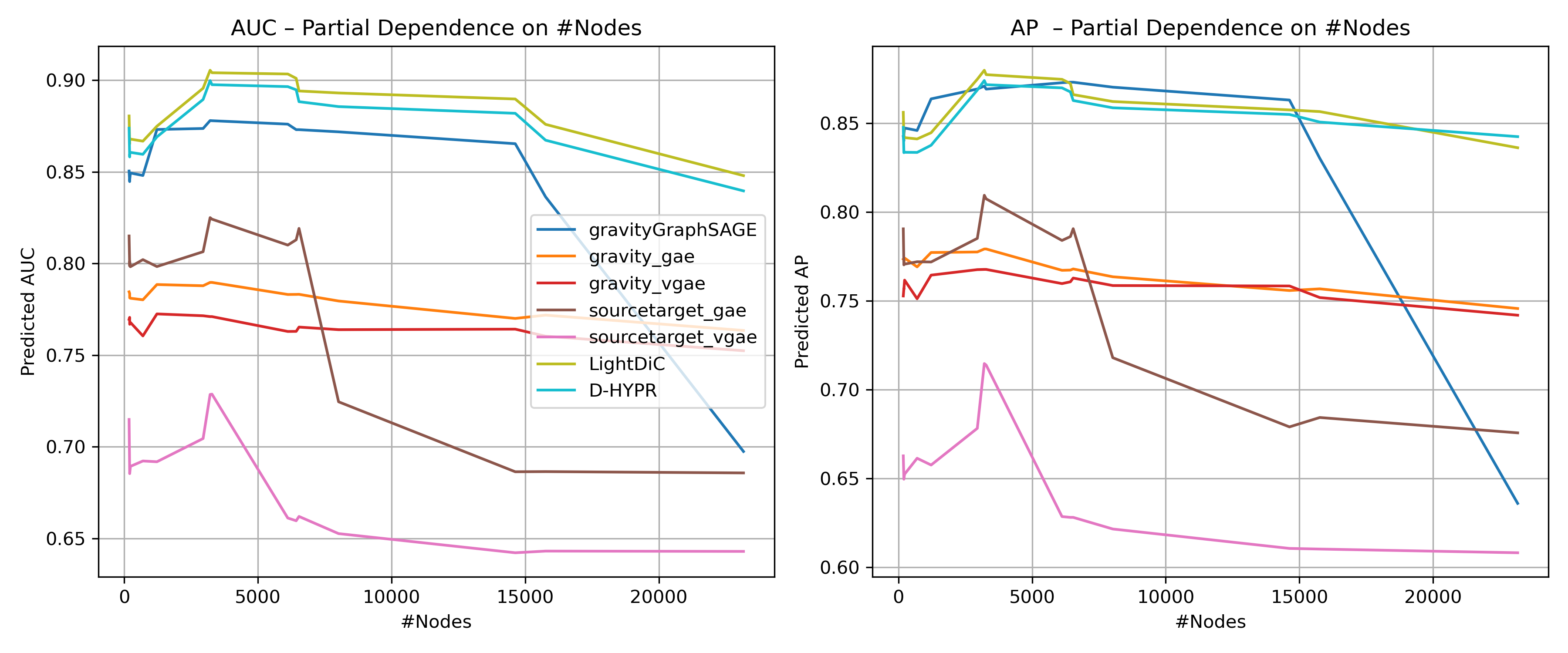}
    \caption{Random Forest Regression Partial Dependence Plot on the number of nodes in graphs.}
    \label{fig:ablation_nodes}
\end{figure}
\begin{figure}[h]
    \centering
    \includegraphics[width=0.9\linewidth]{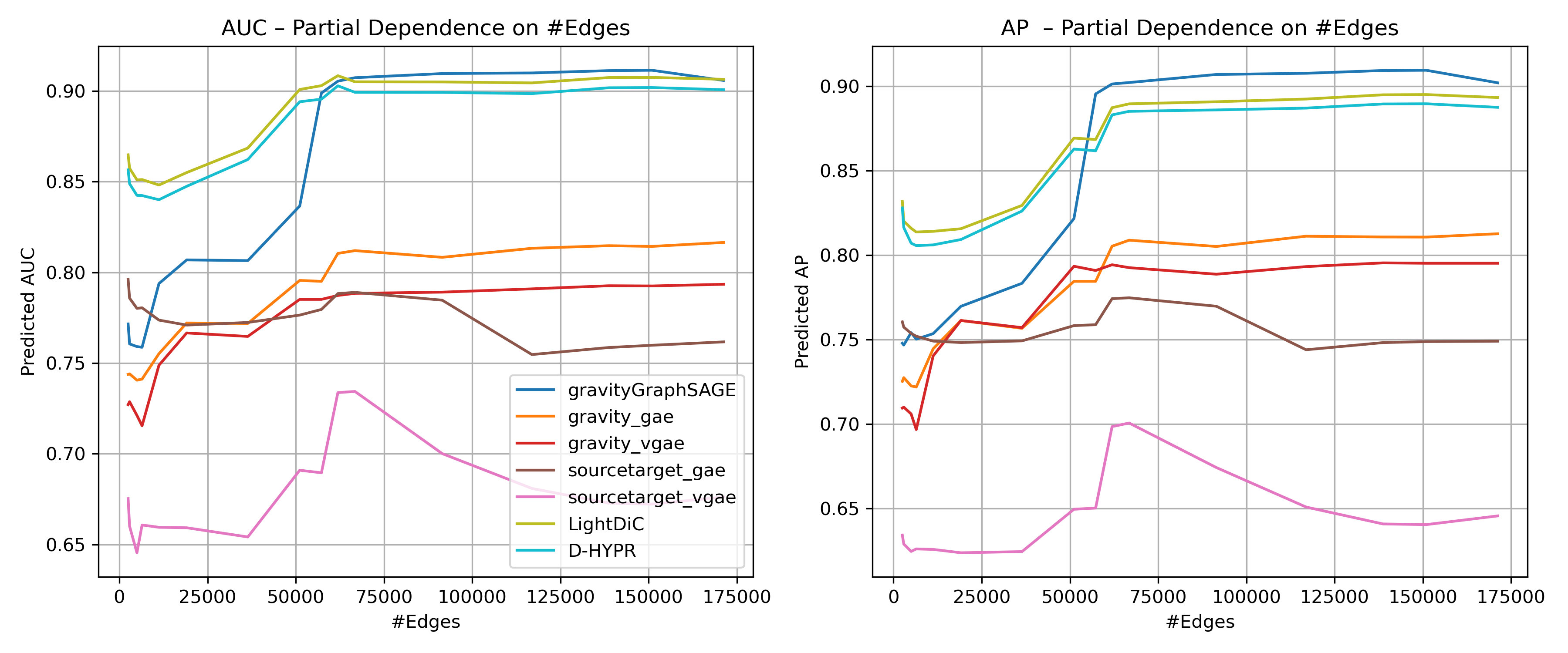}
    \caption{Random Forest Regression Partial Dependence Plot on the number of edges in graphs.}
    \label{fig:ablation_edges}
\end{figure}
\begin{figure}
    \centering
    \includegraphics[width=0.9\linewidth]{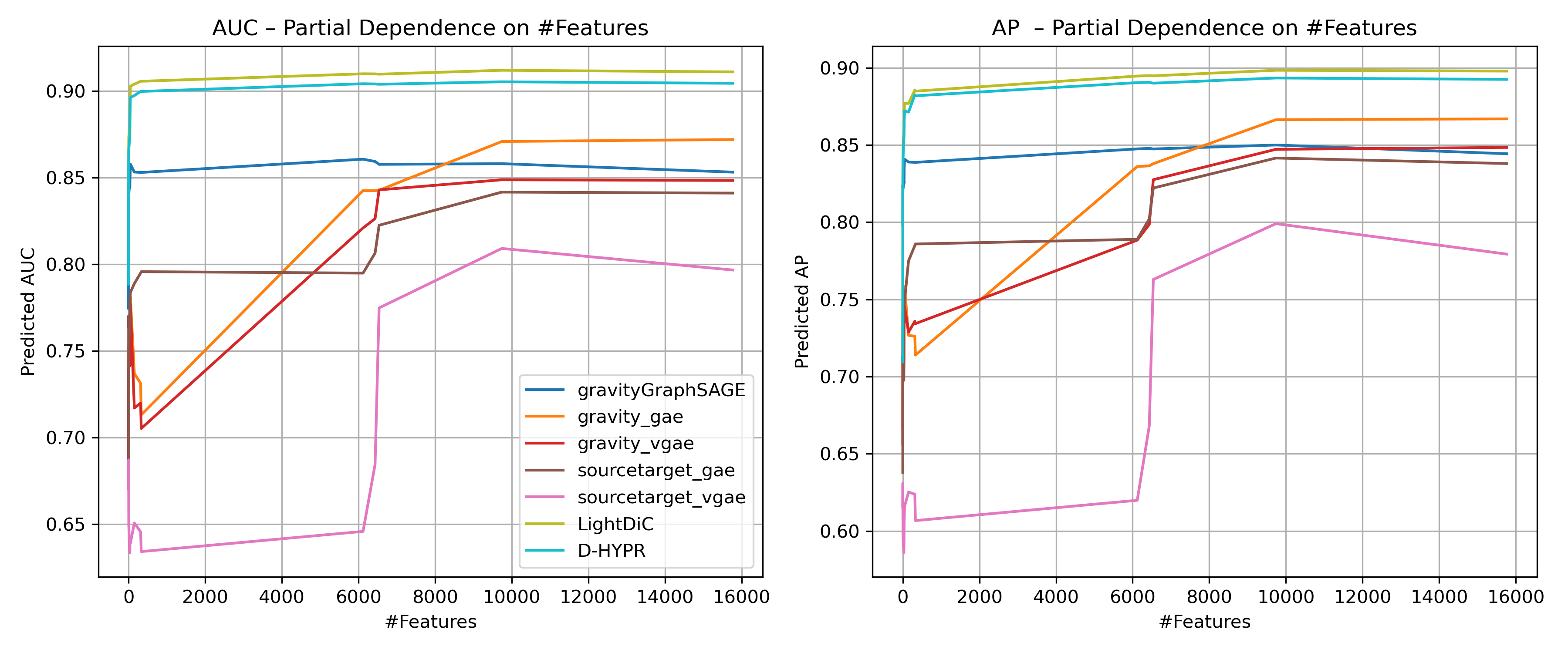}
    \caption{Random Forest Regression Partial Dependence Plot on the number of node features in graphs.}
    \label{fig:ablation_features}
\end{figure}

\begin{table}[t]
    \centering
    \caption{\label{tab:datasets}Dataset Statistics. 
    %Symbol 
    ``*" indicates 
    %that the dataset is a collection of more 
    datasets comprising multiple networks;
    %. In these cases we select the one with maximum 
    for these we select the network with the largest number of nodes.}
    \vskip 0.15in
    \begin{tabular}{cccc}
        \toprule
        \textbf{Dataset} & \textbf{\#Nodes} & \textbf{\#Edges} & \textbf{\#Features} \\
        \midrule
        Cora & \np{2708}   & \np{5429}   & 1433 \\
        Citeseer & \np{3312}  & \np{4591} & \np{3703} \\
        PubMed & \np{19717}  & \np{88651} & \np{500} \\
        \midrule
        foodweb\_little\_rock  & 183   & \np{2494}   & 37   \\
        cintestinalis         & 205   & \np{2903}   & 34   \\
        faculty\_hiring*       & 206   & \np{4988}   & 332  \\
        messal\_shale        & 700   & \np{6444}   & 152  \\
        un\_migrations       & 232   & \np{11228}  & 28   \\
        polblogs             & \np{1224}  & \np{19090}  & 48   \\
        advogato             & \np{6539}  & \np{51127}  & \np{6539}  \\
        inploid              & \np{14629} & \np{57276}  & 6    \\
        faculty\_hiring\_us*  & \np{3284}  & \np{61936}  & 3    \\
        openflights         & \np{3214}  & \np{66771}  & \np{9740} \\
        cora-1998                & \np{23166} & \np{91500}  & 1    \\
        caida\_as*           & \np{8020}  & \np{36406}  & 1    \\
        fly\_larva          & \np{2952}  & \np{116922} & 316  \\
        jung                & \np{6120}  & \np{138706} & \np{6120}  \\
        jdk                 & \np{6434}  & \np{150985} & \np{6434}  \\
        google              & \np{15763} & \np{171206} & \np{15763} \\
        \bottomrule
    \end{tabular}
\end{table}

\begin{table*}[htb]
\centering
\begin{minipage}[t]{0.5\textwidth}
\scalebox{0.7}{%
\begin{tabular}{cccc}
\toprule
\textbf{Dataset} & \textbf{Model} & \textbf{AUC (\%)} & \textbf{AP (\%)} \\
\midrule

\multirow{7}{*}{advogato} & GG-SAGE & 84.11 $\pm$ 3.48 & 84.68 $\pm$ 3.55 \\
\hhline{~---}
& Gravity GAE             & 88.04 $\pm$ 0.24 & 88.79 $\pm$ 0.27 \\
& Gravity VGAE            & 89.81 $\pm$ 0.03 & 90.21 $\pm$ 0.21 \\
& Source/Target GAE       & 91.01 $\pm$ 0.12 & 90.61 $\pm$ 0.23 \\
& Source/Target VGAE      & 87.84 $\pm$ 0.12 & 85.25 $\pm$ 0.19 \\
& LightDiC                & \cellcolor{green}\textbf{91.80 $\pm$ 0.20} & \cellcolor{green}\textbf{91.30 $\pm$ 0.20} \\
& D-HYPR                  & 91.20 $\pm$ 1.00 & 90.70 $\pm$ 1.00 \\
\midrule

\multirow{7}{*}{caida\_as} & GG-SAGE & \cellcolor{green}\textbf{78.56 $\pm$ 3.20} & \cellcolor{green}\textbf{80.74 $\pm$ 2.81} \\
\hhline{~---}
& Gravity GAE             & 60.48 $\pm$ 0.82 & 56.12 $\pm$ 0.73 \\
& Gravity VGAE            & 63.65 $\pm$ 0.91 & 59.47 $\pm$ 0.40 \\
& Source/Target GAE       & 52.02 $\pm$ 2.88 & 50.82 $\pm$ 1.33 \\
& Source/Target VGAE      & 52.38 $\pm$ 6.07 & 50.55 $\pm$ 4.50 \\
& LightDiC                & 78.00 $\pm$ 0.20 & 65.00 $\pm$ 0.20 \\
& D-HYPR                  & 77.40 $\pm$ 1.00 & 64.80 $\pm$ 1.00 \\
\midrule

\multirow{7}{*}{cintestinalis} & GG-SAGE & 70.90 $\pm$ 3.83 & 66.65 $\pm$ 4.02 \\
\hhline{~---}
& Gravity GAE             & 76.63 $\pm$ 0.62 & 75.11 $\pm$ 0.36 \\
& Gravity VGAE            & 75.58 $\pm$ 0.33 & 75.81 $\pm$ 0.43 \\
& Source/Target GAE       & 70.01 $\pm$ 6.02 & 64.57 $\pm$ 6.30 \\
& Source/Target VGAE      & 54.92 $\pm$ 1.03 & 51.10 $\pm$ 1.25 \\
& LightDiC                & \cellcolor{green}\textbf{77.30 $\pm$ 0.20} & \cellcolor{green}\textbf{75.90 $\pm$ 0.20} \\
& D-HYPR                  & 76.40 $\pm$ 1.00 & 75.00 $\pm$ 1.00 \\
\midrule

\multirow{7}{*}{cora} & GG-SAGE & 48.67 $\pm$ 2.18 & 44.44 $\pm$ 0.98 \\
\hhline{~---}
& Gravity GAE             & 64.44 $\pm$ 0.13 & 61.04 $\pm$ 0.13 \\
& Gravity VGAE            & 60.63 $\pm$ 0.46 & 55.11 $\pm$ 0.36 \\
& Source/Target GAE       & 51.40 $\pm$ 1.97 & 47.66 $\pm$ 1.88 \\
& Source/Target VGAE      & 55.52 $\pm$ 2.03 & 51.22 $\pm$ 1.54 \\
& LightDiC                & \cellcolor{green}\textbf{65.00 $\pm$ 0.20} & \cellcolor{green}\textbf{62.60 $\pm$ 0.20} \\
& D-HYPR                  & 64.40 $\pm$ 1.00 & 62.00 $\pm$ 1.00 \\
\midrule

\multirow{7}{*}{faculty\_hiring} & GG-SAGE & 73.77 $\pm$ 1.91 & 77.38 $\pm$ 1.37 \\
\hhline{~---}
& Gravity GAE             & 60.96 $\pm$ 0.43 & 61.61 $\pm$ 0.94 \\
& Gravity VGAE            & 61.45 $\pm$ 0.78 & 64.22 $\pm$ 1.27 \\
& Source/Target GAE       & 83.45 $\pm$ 1.42 & 82.10 $\pm$ 1.33 \\
& Source/Target VGAE      & 52.54 $\pm$ 10.03 & 53.56 $\pm$ 8.89 \\
& LightDiC                & \cellcolor{green}\textbf{84.00 $\pm$ 0.20} & \cellcolor{green}\textbf{82.50 $\pm$ 0.20} \\
& D-HYPR                  & 83.40 $\pm$ 1.00 & 81.90 $\pm$ 1.00 \\
\midrule

\multirow{7}{*}{faculty\_hiring\_us} & GG-SAGE & \textbf{95.93 $\pm$ 0.43} & \cellcolor{green}\textbf{94.95 $\pm$ 0.31} \\
\hhline{~---}
& Gravity GAE             & 83.28 $\pm$ 0.63 & 82.79 $\pm$ 0.65 \\
& Gravity VGAE            & 76.32 $\pm$ 2.60 & 80.17 $\pm$ 1.11 \\
& Source/Target GAE       & 95.97 $\pm$ 0.05 & 93.60 $\pm$ 0.16 \\
& Source/Target VGAE      & 94.31 $\pm$ 0.90 & 89.62 $\pm$ 2.43 \\
& LightDiC                & \cellcolor{green}\textbf{96.40 $\pm$ 0.20} & 94.70 $\pm$ 0.20 \\
& D-HYPR                  & 95.80 $\pm$ 1.00 & 94.20 $\pm$ 1.00 \\
\midrule

\multirow{7}{*}{fly\_larva} & GG-SAGE & \cellcolor{green}\textbf{93.64 $\pm$ 0.56} & \cellcolor{green}\textbf{93.28 $\pm$ 0.56} \\
\hhline{~---}
& Gravity GAE             & 79.59 $\pm$ 0.42 & 79.98 $\pm$ 0.48 \\
& Gravity VGAE            & 79.55 $\pm$ 0.55 & 79.72 $\pm$ 0.50 \\
& Source/Target GAE       & 78.90 $\pm$ 0.79 & 77.43 $\pm$ 0.38 \\
& Source/Target VGAE      & 68.23 $\pm$ 2.90 & 67.47 $\pm$ 3.47 \\
& LightDiC                & 92.50 $\pm$ 0.20 & 92.60 $\pm$ 0.20 \\
& D-HYPR                  & 91.90 $\pm$ 1.00 & 91.90 $\pm$ 1.00 \\
\midrule

\multirow{7}{*}{foodweb\_little\_rock} & GG-SAGE & 83.81 $\pm$ 1.63 & 79.43 $\pm$ 1.70 \\
\hhline{~---}
& Gravity GAE             & 75.86 $\pm$ 5.88 & 69.44 $\pm$ 5.47 \\
& Gravity VGAE            & 72.48 $\pm$ 1.62 & 63.32 $\pm$ 1.11 \\
& Source/Target GAE       & 95.28 $\pm$ 0.16 & 92.81 $\pm$ 0.38 \\
& Source/Target VGAE      & 79.18 $\pm$ 1.30 & 69.13 $\pm$ 1.20 \\
& LightDiC                & \cellcolor{green}\textbf{95.80 $\pm$ 0.20} & \cellcolor{green}\textbf{93.30 $\pm$ 0.20} \\
& D-HYPR                  & 95.40 $\pm$ 1.00 & 92.90 $\pm$ 1.00 \\
\end{tabular}%
}
\end{minipage}\hfill
\begin{minipage}[t]{0.5\textwidth}
\scalebox{0.7}{%
\begin{tabular}{cccc}
\toprule
\textbf{Dataset} & \textbf{Model} & \textbf{AUC (\%)} & \textbf{AP (\%)} \\
\midrule

\multirow{7}{*}{google} & GG-SAGE & 88.79 $\pm$ 1.29 & 89.14 $\pm$ 0.66 \\
\hhline{~---}
& Gravity GAE             & 92.46 $\pm$ 0.52 & 92.49 $\pm$ 0.59 \\
& Gravity VGAE            & 88.22 $\pm$ 0.86 & 90.34 $\pm$ 0.79 \\
& Source/Target GAE       & 81.59 $\pm$ 0.84 & 81.39 $\pm$ 0.60 \\
& Source/Target VGAE      & 72.56 $\pm$ 5.57 & 70.16 $\pm$ 6.11 \\
& LightDiC                & \cellcolor{green}\textbf{93.00 $\pm$ 0.20} & \cellcolor{green}\textbf{93.00 $\pm$ 0.20} \\
& D-HYPR                  & 92.40 $\pm$ 1.00 & 92.50 $\pm$ 1.00 \\
\midrule

\multirow{7}{*}{inploid} & GG-SAGE & \cellcolor{green}\textbf{92.92 $\pm$ 2.05} & \cellcolor{green}\textbf{92.96 $\pm$ 2.30} \\
\hhline{~---}
& Gravity GAE             & 74.48 $\pm$ 0.81 & 73.43 $\pm$ 1.21 \\
& Gravity VGAE            & 77.05 $\pm$ 0.45 & 78.67 $\pm$ 0.59 \\
& Source/Target GAE       & 58.52 $\pm$ 5.00 & 53.07 $\pm$ 4.39 \\
& Source/Target VGAE      & 55.81 $\pm$ 1.62 & 49.85 $\pm$ 1.12 \\
& LightDiC                & 92.00 $\pm$ 0.20 & 88.80 $\pm$ 0.20 \\
& D-HYPR                  & 91.20 $\pm$ 1.00 & 88.20 $\pm$ 1.00 \\
\midrule

\multirow{7}{*}{jdk} & GG-SAGE & \cellcolor{green}\textbf{97.47 $\pm$ 0.38} & \cellcolor{green}\textbf{96.62 $\pm$ 0.67} \\
\hhline{~---}
& Gravity GAE             & 87.59 $\pm$ 0.65 & 87.17 $\pm$ 1.88 \\
& Gravity VGAE            & 84.50 $\pm$ 0.40 & 81.99 $\pm$ 0.26 \\
& Source/Target GAE       & 78.15 $\pm$ 1.37 & 78.87 $\pm$ 1.00 \\
& Source/Target VGAE      & 58.04 $\pm$ 10.99 & 56.28 $\pm$ 13.20 \\
& LightDiC                & 96.50 $\pm$ 0.20 & 96.00 $\pm$ 0.20 \\
& D-HYPR                  & 95.90 $\pm$ 1.00 & 95.50 $\pm$ 1.00 \\
\midrule

\multirow{7}{*}{jung} & GG-SAGE & \cellcolor{green}\textbf{97.03 $\pm$ 1.27} & \cellcolor{green}\textbf{96.07 $\pm$ 1.71} \\
\hhline{~---}
& Gravity GAE             & 87.93 $\pm$ 0.34 & 87.50 $\pm$ 1.55 \\
& Gravity VGAE            & 85.02 $\pm$ 0.18 & 82.48 $\pm$ 0.22 \\
& Source/Target GAE       & 75.88 $\pm$ 2.19 & 77.61 $\pm$ 1.11 \\
& Source/Target VGAE      & 59.51 $\pm$ 10.40 & 57.10 $\pm$ 13.52 \\
& LightDiC                & 96.00 $\pm$ 0.20 & 95.40 $\pm$ 0.20 \\
& D-HYPR                  & 95.60 $\pm$ 1.00 & 95.00 $\pm$ 1.00 \\
\midrule

\multirow{7}{*}{messal\_shale} & GG-SAGE & 73.38 $\pm$ 6.44 & \textbf{74.91 $\pm$ 6.04} \\
\hhline{~---}
& Gravity GAE             & 62.67 $\pm$ 6.78 & 60.43 $\pm$ 4.19 \\
& Gravity VGAE            & 52.57 $\pm$ 0.60 & 54.56 $\pm$ 1.00 \\
& Source/Target GAE       & 84.04 $\pm$ 0.71 & 79.66 $\pm$ 1.06 \\
& Source/Target VGAE      & 69.06 $\pm$ 4.57 & 64.83 $\pm$ 2.50 \\
& LightDiC                & \cellcolor{green}\textbf{84.30 $\pm$ 0.20} & \cellcolor{green}\textbf{79.80 $\pm$ 0.20} \\
& D-HYPR                  & 83.20 $\pm$ 1.00 & 78.70 $\pm$ 1.00 \\
\midrule

\multirow{7}{*}{openflights} & GG-SAGE & \cellcolor{green}\textbf{98.01 $\pm$ 0.51} & \cellcolor{green}\textbf{98.52 $\pm$ 0.26} \\
\hhline{~---}
& Gravity GAE             & 95.09 $\pm$ 0.23 & 94.93 $\pm$ 0.22 \\
& Gravity VGAE            & 93.77 $\pm$ 0.22 & 94.52 $\pm$ 0.15 \\
& Source/Target GAE       & 96.78 $\pm$ 0.11 & 96.53 $\pm$ 0.14 \\
& Source/Target VGAE      & 94.01 $\pm$ 0.56 & 93.29 $\pm$ 0.50 \\
& LightDiC                & 97.50 $\pm$ 0.20 & 97.90 $\pm$ 0.20 \\
& D-HYPR                  & 97.00 $\pm$ 1.00 & 97.40 $\pm$ 1.00 \\
\midrule

\multirow{7}{*}{polblogs} & GG-SAGE & \cellcolor{green}\textbf{90.09 $\pm$ 0.55} & \cellcolor{green}\textbf{83.76 $\pm$ 0.34} \\
\hhline{~---}
& Gravity GAE             & 83.84 $\pm$ 0.28 & 81.77 $\pm$ 0.81 \\
& Gravity VGAE            & 84.14 $\pm$ 0.22 & 83.09 $\pm$ 0.72 \\
& Source/Target GAE       & 77.48 $\pm$ 2.17 & 73.01 $\pm$ 2.16 \\
& Source/Target VGAE      & 60.56 $\pm$ 1.30 & 57.80 $\pm$ 2.72 \\
& LightDiC                & 88.50 $\pm$ 0.20 & 83.10 $\pm$ 0.20 \\
& D-HYPR                  & 87.80 $\pm$ 1.00 & 82.50 $\pm$ 1.00 \\
\midrule

\multirow{7}{*}{un\_migrations} & GG-SAGE & \cellcolor{green}\textbf{82.29 $\pm$ 0.96} & 75.68 $\pm$ 0.92 \\
\hhline{~---}
& Gravity GAE             & 75.97 $\pm$ 0.50 & 77.05 $\pm$ 0.55 \\
& Gravity VGAE            & 75.62 $\pm$ 0.38 & 76.48 $\pm$ 0.63 \\
& Source/Target GAE       & 72.16 $\pm$ 8.47 & 65.61 $\pm$ 8.27 \\
& Source/Target VGAE      & 62.13 $\pm$ 1.17 & 55.25 $\pm$ 1.76 \\
& LightDiC                & 81.80 $\pm$ 0.20 & \cellcolor{green}\textbf{77.40 $\pm$ 0.20} \\
& D-HYPR                  & 81.00 $\pm$ 1.00 & 76.80 $\pm$ 1.00 \\
\end{tabular}%
}
\end{minipage}

\caption{Performance of models on selected Netzschleuder datasets. Best results are higlighted in green and in bold are indicated comparable performances.}
\label{tab:results}
\end{table*}

%\end{document}

\end{document}